\documentclass[conference]{IEEEtran}
\IEEEoverridecommandlockouts
\usepackage[square, numbers, sort&compress]{natbib}
\usepackage{amsmath,amssymb,amsfonts}
\usepackage{amsthm}

\usepackage{algorithm}
\usepackage{algpseudocode}
\usepackage{bm}
\usepackage{multirow}
\usepackage{booktabs}
\usepackage{tabularx}
\usepackage{graphicx}
\usepackage{textcomp}
\usepackage{xcolor}
\usepackage[most]{tcolorbox}
\usepackage{colortbl}
\usepackage{makecell}
\usepackage{url}
\usepackage[hidelinks]{hyperref}
\usepackage{tikz}
\usetikzlibrary{positioning,arrows.meta}

\def\BibTeX{{\rm B\kern-.05em{\sc i\kern-.025em b}\kern-.08em
    T\kern-.1667em\lower.7ex\hbox{E}\kern-.125emX}}

\newif\ifprovisional
\provisionalfalse   % keine \prov-Stellen mehr im Text, Schalter bleibt fuer kuenftige Runden
\ifprovisional
  
\else
  
\fi

\begin{document}

\title{Learning From Static Context in Federated Time Series Forecasting \\
\thanks{
    This work was supported by the German Federal Ministry of Research, Technology and Space under Grant 13FH587KX1 (FederatedForecasts) \\}
}

\author{\IEEEauthorblockN{Viktor Walter}
\IEEEauthorblockA{\textit{Faculty of Management Science and Engineering} \\
\textit{Karlsruhe University of Applied Sciences (HKA)}\\
Karlsruhe, Germany\\
viktor.walter@h-ka.de}
\and
\IEEEauthorblockN{Andreas Wagner}
 \IEEEauthorblockA{\textit{Faculty of Management Science and Engineering} \\
 \textit{Karlsruhe University of Applied Sciences (HKA)}\\
 Karlsruhe, Germany\\
 \textit{Fraunhofer Institute for Industrial Mathematics (ITWM)}\\
 Kaiserslautern, Germany\\
 andreas.wagner@h-ka.de}
\and
\IEEEauthorblockN{Fabian Stricker}
\IEEEauthorblockA{\textit{Institute of Data-Centric
Software Systems (IDSS)} \\
\textit{Karlsruhe University of Applied Sciences (HKA)}\\
Karlsruhe, Germany\\
fabian.stricker@h-ka.de}
}

\maketitle

\begin{abstract}
Time series forecasting is shifting from locally trained, per-series models to global models trained across many heterogeneous series, which requires learning what sets the series apart.
When the raw data cannot be centralized, such a global model is trained with federated learning (FL), one client per data owner.
State-of-the-art forecasters such as the Temporal Fusion Transformer (TFT) tell series apart through a \emph{static} input that describes each series and does not change over time.
In the special case of a federated setting where each client holds a single series, this static input is constant within each client and varies only across clients.
We ask whether a federated TFT can learn from it.
Per-batch FedAdam, which aggregates gradients after every batch, uses the static input almost as well as central training, whereas communication-efficient schemes with local epochs (FedAvg, FedAdam+Prox) extract far less from it and gain nothing significant on either application dataset.
At the same local computation, an Adam-compatible SCAFFOLD keeps the static pathway responsive on wind power and gains significantly there. On building energy its error still drops, but the static pathway falls silent, so drift and communication frequency cannot be fully separated as the cause.
Only per-batch FedAdam matches or beats the per-client local models across all datasets.
We confirm the effect on building-energy meter data, a wind-power dataset, and a controlled synthetic dataset. The loss is not specific to the TFT: on the synthetic and the wind data two further architectures with different static-conditioning mechanisms show it as well.
\end{abstract}


% =====================================================================
% Zahlen aus der Experimentrunde vom 2026-09-23.
%
% ZWECK: Jede Zahl im Manuskript, die nicht in einer Tabelle steht, steht
% hier genau einmal.
%
% STATUS: [F] endgueltig, alle Seeds gerechnet
%
% Quelle der Zahlen: docs/experimente_2026-09-23.md
% =====================================================================

% ---------------------------------------------------------------------
% Relative Sensitivitaet auf der Synthetik, alle Verfahren, TFT
% [F] 5 Seeds, aus outputs/synth/sensitivity_analysis/
% ---------------------------------------------------------------------
\newcommand{\sensSynthCentral}{4.50}        % [F]
\newcommand{\sensSynthFedavgRel}{0.11}      % [F] normiert, roh 0.5027/4.5008
\newcommand{\sensSynthProx}{0.18}           % [F] normiert, roh 0.7958/4.5008
\newcommand{\sensSynthScaffold}{0.16}       % [F] normiert, roh 0.7235/4.5008

% ---------------------------------------------------------------------
% Kopplung nur des statischen Pfades (static-path coupling), BDG2
% Variante A (Juni): temporaler Pfad alle 50 Batches aggregiert (E ~ 0.16),
% fuenf Seeds. Steht im Text nur als Kontrast.
% ---------------------------------------------------------------------
\newcommand{\fedstaticRmse}{0.453}          % [F]
\newcommand{\fedstaticP}{0.024}             % [F]
\newcommand{\staticPathShare}{22.3}         % [F] Anteil an den Parametern
% Variante B (2026-09-24, fsepoch, l1): temporaler Pfad eine volle Epoche
% (314 Batches) mit FedAdam+Prox, fuenf Seeds. logs/fsepoch_table.log,
% outputs/static_analysis_sweep/ Label FedStatic(E=1).
\newcommand{\fsepochRmse}{0.539}            % [F] mit statischem Input
\newcommand{\fsepochRmseNostatic}{0.556}    % [F]
\newcommand{\fsepochP}{0.013}               % [F]
\newcommand{\fsepochSens}{0.015}            % [F] normiert (absolut 0.0009)
\newcommand{\fsepochRounds}{44}             % [F]
\newcommand{\fsepochCommShare}{186}         % [F] Prozent von per-batch (17 553 gegen 9 456 MB/Client)

% ---------------------------------------------------------------------
% Zahl lokaler Schritte (Dosis-Wirkung), BDG2
% [F] vier Seeds 42/123/456/789 (Stand 2026-09-24); Seed 1234 wird
% nachgerechnet, sobald SCAFFOLD die Maschinen freigibt. Vier Seeds sind
% dieselbe Zahl wie in Tabelle VI fuer Central-nostatic und Local, der Text
% weist sie aus. Alte Dreiseedwerte: 0.483/0.571 FedAvg, 0.461/0.594 Prox.
% ---------------------------------------------------------------------
\newcommand{\sweepAvgQuarter}{0.486}        % [F] FedAvg, E = 0.25, 4 Seeds
\newcommand{\sweepAvgTwo}{0.575}            % [F] FedAvg, E = 2
\newcommand{\sweepProxQuarter}{0.465}       % [F] FedAdam+Prox, E = 0.25
\newcommand{\sweepProxTwo}{0.599}           % [F] FedAdam+Prox, E = 2

% ---------------------------------------------------------------------
% Violinabbildung BDG2 (Aggregation wie Tabelle VI: je Seed die Differenz
% zur lokalen Baseline desselben Seeds, dann ueber Seeds mitteln)
% ---------------------------------------------------------------------
\newcommand{\medPerbatchNostatic}{0.046}    % [F]
\newcommand{\medPerbatchStatic}{0.058}      % [F]

% ---------------------------------------------------------------------
% Kontrolle bei gleicher Schrittzahl auf BDG2: FedAdam+Prox E = 0.1 (vier
% Seeds 42/123/456/789) gegen SCAFFOLD (50 Schritte je Runde, fuenf Seeds).
% Quelle: /tmp/e01_table.py auf l1 (make_summary_table) und
% outputs/static_analysis_sweep/importance_results.csv
% ---------------------------------------------------------------------
% Alle Sensitivitaeten im Text sind auf den zentralen TFT desselben
% Datensatzes normiert, wie Abbildung 2 und Tabelle V. Referenz BDG2:
% 0.0619 absolut. Umrechnung im Kommentar jeweils dahinter.
\newcommand{\rmseBdgProxEzeroone}{0.461}    % [F] mit statischem Input, 4 Seeds
\newcommand{\pBdgProxEzeroone}{0.353}       % [F]
\newcommand{\sensBdgProxEzeroone}{0.073}    % [F] 4 Seeds (absolut 0.0045)
% SCAFFOLD mit einer lokalen Epoche auf BDG2 (Neulauf 2026-09-25, l1+l2,
% fuenf Seeds). Ersetzt die 50-Schritt-Zeile der eingereichten Fassung
% (0.468/0.444, p = 1.000, Sensitivitaet 0.327).
\newcommand{\scaffoldEpochRmse}{0.502}      % [F] mit statischem Input, 5 Seeds
\newcommand{\scaffoldEpochRmseNostatic}{0.509} % [F]
\newcommand{\scaffoldEpochP}{0.006}         % [F]
\newcommand{\scaffoldEpochSens}{0.002}      % [F] normiert (absolut 0.0001)

% ---------------------------------------------------------------------
% Kontrolle mit konstantem statischem Input (Laeufe *_ones vom Juni).
% Gleiche Architektur, gleiche Parameterzahl, nur ohne Information.
% Fehler je Gebaeude ueber die gemeinsamen Seeds gemittelt, Wilcoxon ueber
% 32 Gebaeude. per-batch: Seeds 42/123. FedAdam+Prox: Seeds 42/123/456.
% Quelle: ones_check.py auf l1.
% ---------------------------------------------------------------------
\newcommand{\onesProxReal}{0.516}           % [F] echter statischer Input
\newcommand{\onesProxConst}{0.519}          % [F] konstanter Input
\newcommand{\onesProxNone}{0.548}           % [F] ohne statischen Pfad
\newcommand{\onesProxP}{0.70}               % [F] echt gegen konstant
\newcommand{\onesPbReal}{0.375}             % [F]
\newcommand{\onesPbConst}{0.388}            % [F]

% Sensitivitaet je E auf BDG2 (outputs/static_analysis_sweep/, 3 Seeds
% 42/123/456 fuer E != 0.1 und E != 1; E = 1 aus outputs/static_analysis/)
\newcommand{\sensProxEquarter}{0.030}       % [F] E = 0.25, 4 Seeds (absolut 0.0019)
% Normiert, vier Seeds: E = 0.5 misst 0.023, E = 2 misst 0.004, E = 1
% (fuenf Seeds) 0.005 — im Text als "below 0.01".
% FedAvg normiert: 0.001 bei E = 0.25, 0.000 bei E = 1 und E = 2.

% ---------------------------------------------------------------------
% Zahl der Seeds je Datensatz
% ---------------------------------------------------------------------
\newcommand{\seedsBdg}{four to five}        % [F] FL-Verfahren 5, Central-nostatic und Local 4
\newcommand{\seedsWind}{four}               % [F] Seeds 42/123/456/789

\section{Introduction}
\label{sec:intro}

Time series forecasting increasingly relies on large, deep models trained across many heterogeneous series, rather than one specialized model per series.
The data come from diverse sources whose technical characteristics shape each series.
In renewable energy forecasting, for example, technical differences determine the shape of a time series of electricity production and may have to be learned along with it~\cite{walterFederatedLearningBasedForecasting2024}.
% For wind power these include the hub height, rotor diameter, and turbine age, and for photovoltaics the orientation, tilt, and shading of the panels~\cite{walter2025strategies}.
Another domain common in federated forecasting is electricity demand forecasting~\cite{bose2023federated}, where office buildings, hospitals, schools, and homes follow very different patterns.
A model exposed to many series must therefore learn what sets them apart.
Training one global model across many series instead of one model per series is a recent shift~\cite{montero2021principles}, and a federated forecaster is such a global model.

A state-of-the-art forecasting architecture is the Temporal Fusion Transformer (TFT)~\cite{lim2021temporal}.
It uses a dedicated \emph{static} input to tell entities apart.
A static feature is time-invariant and describes the series, in the simplest case as a categorical identifier, or continuously as hub height for wind turbines, floor area for buildings, or tilt angle for photovoltaic systems.

Pooling the raw data is not an option in these domains.
Load profiles are personal data, because non-intrusive load monitoring infers occupancy and appliance use from them~\cite{hart1992nonintrusive}.
Wind and photovoltaic operators treat power curves and supervisory control data as trade secrets, and marketers may not pass on the data of the plants they market~\cite{walterFederatedLearningBasedForecasting2024}.
Federated learning (FL) trains one model across the owners without moving their data, one client per owner.

In the special case where each client holds a single entity, the client boundary is the entity boundary and the static feature is constant within a client.
This case raises the problems we investigate in this paper: a constant feature has no variance, so a client cannot learn from it on its own, and its information lies entirely in how it varies across clients, which only federated training can access.

We study this with a federated TFT trained under several aggregation schemes, a probe that measures static-input usage directly, two diagnostics that localize the loss, and three datasets where the role of the static input is known or controllable.
The answers to the three questions below are the \textbf{contributions} of this paper.

\begin{tcolorbox}[colback=black!3,colframe=black!45,boxrule=0.4pt,
  left=4pt,right=4pt,top=3pt,bottom=3pt,arc=1pt]
\textbf{RQ1} Can a federated TFT learn from a client-constant static input?\\
\emph{Fully only under per-batch FedAdam, which uses it almost as well as central training. Schemes with local epochs extract a fraction at most.}\\[3pt]
\textbf{RQ2} Is the limit the communication frequency or the client drift?\\
\emph{Not conclusively. Fewer local steps help, but they raise the communication frequency and lower the drift together. Drift correction at a fixed frequency, the one experiment that separates the two, keeps part of the static pathway on wind power and gains significantly there, but recovers little on the synthetic data and nothing on building energy.}\\[3pt]
\textbf{RQ3} Where does the failure sit, and what does it cost to avoid?\\
\emph{Not in the static encoder alone. On building energy, the loss grows with the number of local steps under a proximal term and is complete under FedAvg, and coupling only the static encoder per batch leaves the static pathway silent. Avoiding it costs one communication round per gradient step.}
\end{tcolorbox}

\section{Related Work}
\label{sec:related_work}

Federated learning trains a shared model across clients without sharing raw data~\cite{mcmahan2017communication}.
It is widely used for energy time series forecasting, where data are private and siloed~\cite{walterFederatedLearningBasedForecasting2024,bose2023federated}, and its recurring obstacle is client heterogeneity.
Domain-informed features are engineered to reduce input heterogeneity across clients, so that a single shared model fits them more easily~\cite{walter2025strategies}.
Another mitigation is personalization, which keeps part of the model local to each client~\cite{arivazhagan2019federated}.
We exclude personalization and keep a single global model.
Personalization holds the last layers local and aggregates the rest, whereas the static input enters at the start of the network, and personalizing the static pathway would defeat its purpose, because its input is constant within a client.

Deep forecasting models separate time-varying inputs from static inputs that describe each series.
The Temporal Fusion Transformer encodes static covariates into context vectors that initialize the states of its long short-term memory (LSTM) encoder and enrich the temporal features~\cite{lim2021temporal}.
Global models trained over many series rely on static entity embeddings to tell series apart~\cite{wang2025qkcv}.
All of this is centralized, where the static input varies across the dataset. None addresses the federated case, where it is constant within a client.
Recent work deploys the federated TFT for battery state-of-health prediction~\cite{lella2025federated} and cross-silo psychiatric risk forecasting~\cite{ahamed2026fedtft}, but does not ask whether the static input is learned under aggregation.

With heterogeneous data, local updates move toward local optima, a phenomenon known as client drift that grows with the number of local steps~\cite{karimireddy2020scaffold}.
The theory of local versus minibatch stochastic gradient descent (SGD) suggests that, in sufficiently heterogeneous regimes, averaging locally updated weights can be worse than aggregating gradients~\cite{woodworth2020minibatch}.
Several methods counteract drift, through control variates~\cite{karimireddy2020scaffold,karimireddy2021breaking}, proximal terms~\cite{li2020federated}, or adaptive server optimization~\cite{reddi2020adaptive}.
These methods concern the convergence of the model as a whole. Layer-wise studies measure which layers diverge most, on image classification~\cite{zhao2018federated,lee2023layerwise}.
None localizes the failure to a functional input pathway such as a static covariate encoder, or uses selective aggregation as a diagnostic.
Bao et al.\ prove that local steps help to learn client-specific features that are discriminative within a client~\cite{bao2024provable}.
A client-constant static input is the opposite case: locally it is indistinguishable from a bias, and only the aggregation across clients can identify it.

\section{Methodology}
\label{sec:methodology}

\subsection{Temporal Fusion Transformer}
\label{subsec:tft}

The TFT is an attention-based model for multi-horizon forecasting~\cite{lim2021temporal}.
It separates two kinds of inputs.
Time-varying inputs are the observed past and the known future covariates, each passed through a variable selection network (VSN), an LSTM encoder--decoder, and a self-attention layer.
A VSN weights the input variables at each time step and keeps the informative ones.
A gated residual network (GRN) is the recurring building block: a gated nonlinear transform with a skip connection.

\begin{figure}[t]
\centering
\includegraphics[width=\columnwidth]{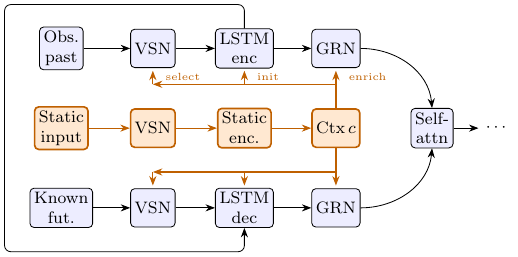}
\caption{TFT with its temporal pathway (blue) and static pathway (orange). Encoder (observed past) and decoder (known future) inputs are separate. The static context vectors condition both at three points: variable selection, LSTM state initialization, and GRN enrichment. Both GRNs feed the shared self-attention}
\label{fig:tft}
\end{figure}

A static covariate encoder maps the static input to four context vectors~\cite{lim2021temporal}.
As shown in Figure~\ref{fig:tft}, they condition three points of the temporal pathway: they enter the variable selection, initialize the hidden and cell states of the LSTM, and enrich the features through a GRN (static enrichment).
%Without the static input, all of these reduce to entity-independent constants, so the model can no longer distinguish entities.

\subsection{Federated Aggregation Schemes}
\label{subsec:aggregation}

We train the TFT across $K$ clients that exchange only model updates.
We compare four aggregation schemes that differ in what they aggregate and how they handle client drift (Table~\ref{tab:schemes}).

\textbf{Per-batch FedAdam} aggregates gradients rather than weights.
Clients compute gradients on a shared model after every batch, the server averages them and applies an Adam step, so coupling is maximal: one aggregation per gradient step.
This is FedAdam~\cite{reddi2020adaptive} in the single-local-step regime, where the client update reduces to the minibatch gradient~\cite{mcmahan2017communication}.

\textbf{FedAvg} aggregates weights.
Each client runs one local epoch, and the server averages the resulting weights, weighted by each client's training sample size~\cite{mcmahan2017communication}.

\textbf{FedAdam+Prox} is a hybrid of two state-of-the-art schemes.
Clients run local epochs with a proximal term that penalizes drift from the global model (FedProx~\cite{li2020federated}).
The server treats the averaged weight deltas as a pseudo-gradient and applies an adaptive Adam update (FedAdam~\cite{reddi2020adaptive}).
We use this hybrid as a strong representative of communication-efficient FL, since it combines the two dominant drift remedies~\cite{nguyen2022novel}.

\textbf{SCAFFOLD} adds client- and server-side control variates to FedAvg to correct client drift~\cite{karimireddy2020scaffold}.
We include it as the canonical drift-correction scheme, to test whether explicit drift correction restores the static pathway under local epochs.
Canonical SCAFFOLD assumes SGD client optimizers. Transformer-based models such as our TFT, however, do not train stably under SGD and diverge~\cite{zhang2020adaptive}, so we train all schemes with Adam.
We therefore use an Adam-compatible variant in the style of FAdamGC~\cite{chen2025fadamgc}.
The drift correction is added to the raw gradient before Adam's moment estimation, because applying it afterwards fails through the nonlinear coupling of the two moments.
The control variates use the optimizer-agnostic mean of the raw local gradients rather than the SGD-specific update rule.
Mime~\cite{karimireddy2021breaking} instead carries server-side Adam statistics into every local step.
We keep the structure of SCAFFOLD, whose only change against FedAvg is the drift correction.
This restores stable training, with errors in the range of FedAvg and FedAdam+Prox.
All three schemes with local epochs run one local epoch per round.

\begin{table}[t]
\centering
\caption{Federated aggregation schemes. $b$ is the number of batches in one local epoch: 214 to 314 on building energy, 45 on wind power, 7 on the synthetic data. CV: control variates}
\label{tab:schemes}
\resizebox{\columnwidth}{!}{%
\begin{tabular}{lccc}
\toprule
Scheme & Local steps per round & Aggregated & Server opt. \\
\midrule
Per-batch FedAdam        & 1               & gradients      & Adam \\
FedAvg                   & one epoch ($b$) & weights        & none \\
FedAdam+Prox             & one epoch ($b$) & deltas         & Adam \\
SCAFFOLD                 & one epoch ($b$) & deltas $+$ CV  & none \\
\bottomrule
\end{tabular}}
\end{table}

FedAdam+Prox uses $\beta_1 = 0.9$ and $\beta_2 = 0.99$ on the server, an adaptivity of $\tau = 10^{-3}$, and a proximal weight of $\lambda = 10^{-2}$.

A static feature needs one more federated step before training.
Because it is constant within a client, standardizing it locally would map it to zero, as it has no within-client variance.
We therefore standardize the static features with statistics aggregated globally across clients~\cite{xu2026fedps}, which preserves their cross-client variation.
For each static feature $v$, each client reports its sample count and the sums of $v$ and $v^2$, the server returns the global mean $\mu$ and variance $\sigma^2$, and each client applies $(v-\mu)/\sigma$. This needs one data pass and one communication round.

This step carries a privacy cost.
Because the feature is constant within a client, the reported sum and count already determine its value, so the server learns each client's static feature exactly. Aggregating these statistics relaxes the federated privacy guarantee for the very input we study.
We accept this as a limitation here, since our question is whether the static input is learned, not how to protect it, and we standardize identically across all schemes to keep the comparison fair.
Domain knowledge could set these constants without the exchange, since typical hub heights, rotor diameters, or floor areas are not secret (Section~\ref{sec:conclusion}).

\subsection{Probing Static-Input Usage}
\label{subsec:probes}

Error metrics do not show whether a model uses the static input, so we probe it directly at test time, without retraining.
Let $f_\theta(\mathbf{x}_i, \mathbf{u}_i, s_i)$ be the prediction for test sample $i$ with target $y_i$, from observed inputs $\mathbf{x}_i$, known inputs $\mathbf{u}_i$, and static input $s_i$. Without the index, $\mathbf{x}$, $\mathbf{u}$, $s$, and $y$ collect all test samples.

We measure \emph{static-input sensitivity} as the relative increase in error when the static input is randomly permuted across the test set, the permutation feature importance of Breiman~\cite{breiman2001random}.
We pool the test samples of all clients, evaluate the root mean square error (RMSE) with the true static input, then permute the static input across the pooled set and re-evaluate:
\begin{equation}
S = \frac{\mathrm{RMSE}\big(f_\theta(\mathbf{x},\mathbf{u},\pi(s)),\, y\big)}{\mathrm{RMSE}\big(f_\theta(\mathbf{x},\mathbf{u},s),\, y\big)} - 1,
\end{equation}
where $\pi(s)$ is a random permutation of the static inputs over the pooled test set.
$S$ is zero for a model that ignores the static input and grows with its reliance on it, since shuffling an unused input cannot change the error.
Dividing by the unpermuted error makes $S$ comparable across schemes whose error levels differ.
We report every sensitivity as $\tilde S = S / S_{\text{central}}$, relative to the central TFT on the same dataset, so the scales are comparable across datasets.

\subsection{Two Diagnostics on the Static Pathway}
\label{subsec:localize}

The probe shows whether a model responds to the static input, not what removes the response.
Two controlled variations address that.

The first varies the amount of local computation. We sweep the number of local epochs per round from $E = 0.1$ to $E = 2$ for FedAdam+Prox and from $E = 0.25$ to $E = 2$ for FedAvg, and hold everything else fixed. If local training removes the static input, its sensitivity $\tilde S$ should fall with $E$ rather than vanish at a threshold.

The second isolates the static pathway. We aggregate the parameters of the static covariate encoder after every batch while the temporal pathway keeps its local epochs, and call this \emph{static-path coupling}. The static encoder holds \staticPathShare\,\% of the parameters.

\subsection{Statistical Significance}
\label{subsec:significance}

For every client $k$ we average the test RMSE over seeds, with and without the static input, and write the two means $e_k$ and $e_k^{0}$.
The client is the unit we generalize over, and averaging over seeds first keeps the runs of one seed from counting as independent observations.
We test whether the static input lowers the error with a one-sided Wilcoxon signed-rank test on the $K$ paired differences $e_k^{0} - e_k$ and report the $p$-value, with $p<0.05$ as the threshold.

\section{Experimental Setup}
\label{sec:experimental}

We study the question on two application domains, building energy and wind power, and on controlled synthetic data, where the role of the static feature is known.

\subsection{Building Energy Data}
\label{subsec:bdg2}

We use the Building Data Genome 2 (BDG2) corpus of hourly electricity meter readings~\cite{miller2020building}.
We treat $K=32$ buildings as clients.
Each building is described by its primary use, floor area, year built, and floor count as static features, and by past load, local weather, and calendar features as time-varying inputs (Table~\ref{tab:features}).
The primary use covers categories such as education, residential, office, and healthcare. It is label-encoded and embedded linearly like the other static features.
We use a moving window that advances one time step, with the lookback and horizon of Table~\ref{tab:hyperparams}.

We drop any window that contains a missing value rather than impute.
Up to $20\%$ of the values per client are missing, mostly in contiguous stretches of several days, so the share of dropped windows stays close to the share of missing values.

\subsection{Wind Power Data}
\label{subsec:wind}

As a second domain we consider wind power forecasting~\cite{walter2026wind}, where each of the $K=32$ wind turbines is one client.
The power output is physically simulated and validated against real wind turbines~\cite{walter6346310measurement}, whereas the weather inputs are real numerical weather prediction (NWP) output.
The static features capture the physical configuration that shapes the power output (Table~\ref{tab:features}).
The manufacturer's power curve enters through its cut-in, rated, and cut-out wind speeds.
The time-varying inputs are the lagged power and NWP wind speed.
We draw the NWP wind speed from two systems of different spatial resolution, the $2.2$\,km ICON-D2 of the German Weather Service (DWD) and the high-resolution forecast (HRES) of the European Centre for Medium-Range Weather Forecasts (ECMWF), which we use on a $0.25^\circ$ grid (about $28$\,km, natively $0.1^\circ$ or $9$\,km).
ICON-D2 reaches $48$ hours ahead.
The series run from August 2023 to October 2024.
We use the same moving-window protocol as for the load setting.

\begin{table}[t]
\centering
\caption{Input variables per setup}
\label{tab:features}
\resizebox{\columnwidth}{!}{%
\begin{tabular}{llll}
\toprule
Setup & Observed & Known & Static \\
\midrule
Synthetic & \makecell[l]{target\\ lagged} & \makecell[l]{sin/cos 24h,\\ sin/cos 168h} & \makecell[l]{entity\\ feature} \\
\midrule
\makecell[l]{Building\\ energy} & \makecell[l]{load\\ lagged,\\ air temp.,\\ wind speed} & \makecell[l]{hour,\\ day of week\\ (sin/cos),\\ weekend flag} & \makecell[l]{primary use\\ floor area\\ year built\\ floor count} \\
\midrule
\makecell[l]{Wind\\ power} & \makecell[l]{power\\ lagged} & \makecell[l]{wind speed\\ ICON-D2:\\ 78m, 127m, 184m\\ ECMWF HRES:\\ 100m} & \makecell[l]{turbine age\\ power curve\\ hub height\\ rotor diam.\\ altitude} \\
\bottomrule
\end{tabular}}
\end{table}

\subsection{Synthetic Data}
\label{subsec:data}

The synthetic generator isolates the static pathway.
We generate $K=12$ clients, one per entity.
Each entity has a univariate series that shares a common temporal shape but differs in level $\ell_k$.
The shape is the sum of a daily and a weekly sinusoid.
We add to each entity a static feature, a standardized scalar that is constant over time and unique to the entity, and set the level so that it can be recovered only from this feature.
We mean-center each observed window, so the level cannot be read off the recent history and must come from the static feature, whereas the forecast target keeps its absolute level.
This is a deliberate worst case: real static descriptors such as hub height vary smoothly across entities and generalize, whereas here the static input is used purely as an entity cue.
We add Gaussian observation noise with a standard deviation of $0.35$, against sinusoid amplitudes of $2.0$ and $1.0$, with an independent realization per entity.
We use the same moving-window protocol as for the application domains, with the lookback and horizon of Table~\ref{tab:hyperparams}.
To mimic the data loss of the real setting, we randomly drop up to $20\%$ of the windows per client, comparable to BDG2.

\subsection{Models and Baselines}
\label{subsec:baselines}

We use the TFT of Section~\ref{subsec:tft} and train it with the four federated schemes of Section~\ref{subsec:aggregation}.
Its architecture and the windowing per setup are summarized in Table~\ref{tab:hyperparams}.
All federated schemes use a batch size of 32 and a client learning rate of $10^{-3}$, and the server optimizers a learning rate of $10^{-2}$.
Central training uses a batch size of 256 on building energy, 64 on wind power and 32 on the synthetic data. All models are trained with the pinball loss at the median ($q = 0.5$).

\begin{table}[t]
\centering
\caption{Model architecture and windowing per setup}
\label{tab:hyperparams}
\begin{tabular}{lccc}
\toprule
 & Synthetic & \makecell{Building\\ energy} & \makecell{Wind\\ power} \\
\midrule
$d_{\text{model}}$    & 16 & 64 & 64 \\
Attention heads       & 2  & 4  & 4  \\
Lookback              & 24 & 72 & 48 \\
Horizon               & 24 & 24 & 48 \\
Dropout               & 0 & 0.1 & 0.1 \\
\bottomrule
\end{tabular}
\end{table}
We add two baselines.
\emph{Central} pools all data and trains a single model, with and without the static input.
\emph{Local} trains a separate model per client without the static input, which is meaningless on a single client.

\subsection{Protocol}
\label{subsec:protocol}

We split the building-energy and synthetic series $60/40$ in time into a training and a test fold.
The wind series use their last three months, August to October 2024, as the test fold after twelve months of training, so the wind test fold covers one season.
In all cases the test data are strictly later than the training data.
To set the number of training epochs we hold out the last $20\%$ of the training fold as a validation set and pick the epoch that minimizes the validation RMSE. In federated training this validation RMSE is aggregated across clients, and training stops after 10, 20 and 15 rounds without improvement on building energy, wind power and the synthetic data. For per-batch FedAdam a round is a block of at most 50 gradient steps.
We then retrain on the full training fold for that many epochs and evaluate once on the test fold.
This rules out comparing under-trained models while keeping the test fold untouched during training.
We repeat every configuration over five random seeds and report the mean.
We evaluate with RMSE, mean absolute error (MAE), and $R^2$, the static-input probe of Section~\ref{subsec:probes}, and the significance test of Section~\ref{subsec:significance}. Errors are reported on the log-transformed load standardized per building, on power normalized by rated capacity, and on the raw synthetic scale.

\section{Results}
\label{sec:results}

\subsection{Static-Input Usage Under Federated Aggregation}
\label{subsec:res_synthetic}

Table~\ref{tab:synthetic} reports the synthetic results.
By construction, the static input is the only feature from which the model can recover the level.
Central training and per-batch FedAdam use it almost fully, raising $R^2$ from $0.22$ to $0.95$, on par with the local baseline. The local-epoch schemes fall far short.
All three local-epoch schemes gain significantly but only partially ($R^2$ $0.53$ to $0.62$ against $0.95$, $p \le 0.039$).
SCAFFOLD lies between FedAvg and FedAdam+Prox, so its drift correction adds little on this dataset.
The static input is highly informative, yet only frequent coupling lets the model exploit it fully.

\begin{table}[t]
\centering
\caption{Synthetic data. Mean over seeds and clients. $p$: one-sided Wilcoxon signed-rank test over per-client RMSE pairs. * and shading: $p<0.05$}
\label{tab:synthetic}
\resizebox{\columnwidth}{!}{%
\begin{tabular}{lccccc}
\toprule
Scheme & Static & RMSE & MAE & $R^2$ & $p(\text{static}{\downarrow})$ \\
\midrule
Central               & \checkmark & 0.36 & 0.28 & 0.95 & \cellcolor{green!25}$<$.001* \\
Central               & --         & 1.42 & 1.25 & 0.22 & --     \\
Per-batch FedAdam      & \checkmark & 0.35 & 0.28 & 0.95 & \cellcolor{green!25}$<$.001* \\
Per-batch FedAdam      & --         & 1.43 & 1.26 & 0.22 & --     \\
FedAdam+Prox          & \checkmark & 1.00 & 0.75 & 0.62 & \cellcolor{green!25}0.026* \\
FedAdam+Prox          & --         & 1.43 & 1.25 & 0.21 & --     \\
FedAvg                & \checkmark & 1.11 & 0.87 & 0.53 & \cellcolor{green!25}0.039* \\
FedAvg                & --         & 1.45 & 1.24 & 0.19 & --     \\
SCAFFOLD              & \checkmark & 1.02 & 0.86 & 0.60 & \cellcolor{green!25}0.039* \\
SCAFFOLD              & --         & 1.44 & 1.26 & 0.20 & --     \\
Local                 & --         & 0.36 & 0.29 & 0.95 & --     \\
\bottomrule
\end{tabular}}
\end{table}

\subsection{Probing the Static Pathway}
\label{subsec:res_probes}

\begin{figure}[t]
\centering
\includegraphics[width=\columnwidth]{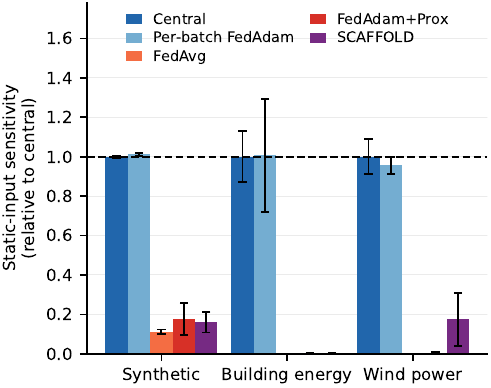}
\caption{Static-input sensitivity relative to central per dataset. Central and per-batch FedAdam stay sensitive on all three datasets. The schemes with local epochs keep a small share on the synthetic data and drop to near zero on the application datasets, except SCAFFOLD on wind power}
\label{fig:sensitivity}
\end{figure}

The static-input sensitivity of Section~\ref{subsec:probes} shows whether the input is used (Figure~\ref{fig:sensitivity}).
Central training and per-batch FedAdam are by far the most sensitive to the static input on all three datasets.
On building energy and wind power, FedAvg and FedAdam+Prox are not: permuting the static input barely changes their error.
On the synthetic data all three local-epoch schemes still respond, but far below central training (\sensSynthFedavgRel{} for FedAvg, \sensSynthScaffold{} for SCAFFOLD and \sensSynthProx{} for FedAdam+Prox), which matches the partial benefit they extract there.
SCAFFOLD stands out from FedAvg and FedAdam+Prox only on wind power, where it retains part of its sensitivity and scatters strongly across seeds.

\subsection{Other Conditioning Mechanisms}
\label{subsec:res_arch}

The static conditioning of the TFT runs through context vectors.
We repeat the experiment with two further architectures that condition differently, a multilayer perceptron (MLP) that receives the static input in the first layer of its shared representation, and a gated recurrent unit (GRU) whose encoder state is initialized and whose decoder state is shifted by it.
Table~\ref{tab:arch} reports the static-input sensitivity of all three under per-batch FedAdam and under FedAvg, on the scale of Figure~\ref{fig:sensitivity}.

On the synthetic data all three architectures lose almost all of their sensitivity under FedAvg, and on wind power the TFT and the GRU lose it entirely while the MLP keeps 0.08 of 0.23. The loss is therefore not a property of the context vectors of the TFT.

On building energy, where the load history already identifies the entity, the TFT loses its response entirely, the MLP has almost none to lose, and the GRU keeps a third of its sensitivity, 0.37 of 1.21.
Both use the hidden size and training setup of the TFT and have 7 to 18 times fewer parameters, so this comparison supports whether the loss occurs, not which architecture forecasts better.

\begin{table}[t]
\centering
\caption{Static-input sensitivity per architecture under per-batch FedAdam (PB) and under FedAvg (FA), relative to the central TFT on the same dataset as in Figure~\ref{fig:sensitivity}. A drop to near zero means the architecture stops responding to the static input}
\label{tab:arch}
\begin{tabular}{lcccccc}
\toprule
 & \multicolumn{2}{c}{Synthetic} & \multicolumn{2}{c}{Building energy} & \multicolumn{2}{c}{Wind power} \\
\cmidrule(lr){2-3}\cmidrule(lr){4-5}\cmidrule(lr){6-7}
 & PB & FA & PB & FA & PB & FA \\
\midrule
TFT & 1.01 & 0.11 & 1.01 & 0.00 & 0.96 & 0.00 \\
MLP & 0.17 & 0.01 & 0.09 & 0.00 & 0.23 & 0.08 \\
GRU & 0.68 & 0.01 & 1.21 & 0.37 & 0.75 & 0.00 \\
\bottomrule
\end{tabular}
\end{table}

\subsection{Building Energy}
\label{subsec:res_bdg2}

Table~\ref{tab:bdg2} shows a similar, though weaker, pattern on BDG2.
Central training and per-batch FedAdam still gain a significant benefit from the static input, but a small one.
FedAvg and FedAdam+Prox gain nothing significant, and SCAFFOLD only appears to.

SCAFFOLD splits the two measurements apart. It reaches RMSE \scaffoldEpochRmse{} with the static input against \scaffoldEpochRmseNostatic{} without, a significant contrast ($p = \scaffoldEpochP{}$), yet its sensitivity is \scaffoldEpochSens{}, as silent as FedAvg.
The model without the static input has no static encoder at all, so the significance test compares two architectures rather than two information states. A constant static input keeps the architecture and removes only the information. In a separate series of three seeds under FedAdam+Prox with one local epoch, the constant matches the real input (\onesProxConst{} against \onesProxReal{}, $p = \onesProxP{}$), whereas under per-batch FedAdam the real input wins (\onesPbReal{} against \onesPbConst{}, $p<0.001$, two seeds). Where the probe is silent, the error difference comes from the additional parameters of the static encoder, which act as learned constants, and not from the entity information.
At equal local computation the drift correction recovers nothing here. Shortening the local run keeps only a trace: FedAdam+Prox at $E = 0.1$ retains a sensitivity of \sensBdgProxEzeroone{} and gains nothing significant (RMSE \rmseBdgProxEzeroone{}, $p = \pBdgProxEzeroone{}$).
On wind power the same correction keeps the static pathway responsive and gains significantly (Section~\ref{subsec:res_wind}).

\begin{table}[t]
\centering
\caption{BDG2 building-energy data. Mean over seeds, $p$ as in Table~\ref{tab:synthetic}. A significant $p$ says the error drops, Figure~\ref{fig:sensitivity} says whether the static input is used}
\label{tab:bdg2}
\resizebox{\columnwidth}{!}{%
\begin{tabular}{lccccc}
\toprule
Scheme & Static & RMSE & MAE & $R^2$ & $p(\text{static}{\downarrow})$ \\
\midrule
Central               & \checkmark & 0.382 & 0.222 & 0.883 & \cellcolor{green!25}$<$.001* \\
Central               & --         & 0.394 & 0.228 & 0.876 & --       \\
Per-batch FedAdam      & \checkmark & 0.382 & 0.220 & 0.880 & \cellcolor{green!25}0.007*   \\
Per-batch FedAdam      & --         & 0.390 & 0.225 & 0.875 & --       \\
FedAdam+Prox          & \checkmark & 0.505 & 0.348 & 0.796 & 0.640    \\
FedAdam+Prox          & --         & 0.511 & 0.349 & 0.797 & --       \\
FedAvg                & \checkmark & 0.521 & 0.361 & 0.791 & 0.865    \\
FedAvg                & --         & 0.519 & 0.356 & 0.796 & --       \\
SCAFFOLD              & \checkmark & 0.502 & 0.347 & 0.810 & \cellcolor{green!25}0.006*   \\
SCAFFOLD              & --         & 0.509 & 0.353 & 0.805 & --       \\
Local                 & --         & 0.513 & 0.298 & 0.802 & --       \\
\bottomrule
\end{tabular}}
\end{table}

\begin{figure}[t]
\centering
\includegraphics[width=\columnwidth]{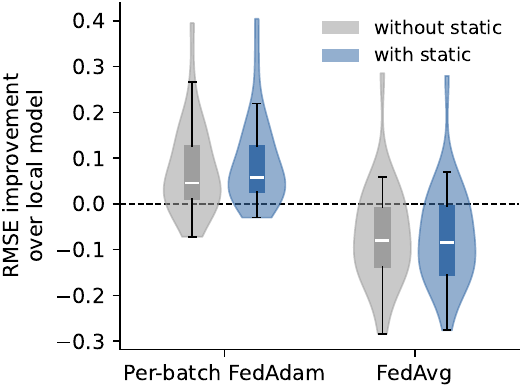}
\caption{Per-building RMSE improvement over the local baseline on building energy, without and with static features. Boxes mark the quartiles and the median. Values above zero beat the local model}
\label{fig:violin}
\end{figure}

Figure~\ref{fig:violin} breaks this down per building. Under per-batch FedAdam the static features move four buildings from worse to better (7 to 3 of 32 worse) and lift the median from \medPerbatchNostatic{} to \medPerbatchStatic{}. Under FedAvg the distribution does not move, and 26 of 32 buildings stay worse than local in both cases.

\subsection{Wind Power}
\label{subsec:res_wind}

Table~\ref{tab:wind} shows the same pattern on wind power, with one informative exception.
Central and per-batch FedAdam gain significantly from the static input, raising $R^2$ from about $0.73$ to $0.76$.
FedAvg and FedAdam+Prox do not ($p>0.1$), and they also reach a lower $R^2$ overall.
SCAFFOLD is the only local-epoch scheme that gains significantly from the static input here ($R^2\,0.63\to0.67$, $p=0.003$), and unlike on building energy the probe agrees: it keeps a sensitivity of 0.18 (Figure~\ref{fig:sensitivity}).
It takes about as many local steps per round as FedAvg, so the difference is its drift correction, which partly recovers the static pathway, though it still trails central, per-batch, and the local baseline.
The local baseline is strong here ($R^2\,0.75$): FedAvg and FedAdam+Prox fall below it, central and per-batch FedAdam exceed it by using the static input.

\begin{table}[t]
\centering
\caption{Wind power data. Mean over seeds, $p$ as in Table~\ref{tab:synthetic}}
\label{tab:wind}
\resizebox{\columnwidth}{!}{%
\begin{tabular}{lccccc}
\toprule
Scheme & Static & RMSE & MAE & $R^2$ & $p(\text{static}{\downarrow})$ \\
\midrule
Central               & \checkmark & 0.094 & 0.055 & 0.761 & \cellcolor{green!25}$<$.001* \\
Central               & --         & 0.100 & 0.059 & 0.725 & --     \\
Per-batch FedAdam      & \checkmark & 0.095 & 0.056 & 0.761 & \cellcolor{green!25}$<$.001* \\
Per-batch FedAdam      & --        & 0.100 & 0.059 & 0.728 & --     \\
FedAdam+Prox          & \checkmark & 0.113 & 0.069 & 0.598 & 0.789  \\
FedAdam+Prox          & --         & 0.113 & 0.068 & 0.601 & --     \\
FedAvg                & \checkmark & 0.112 & 0.067 & 0.627 & 0.610  \\
FedAvg                & --         & 0.112 & 0.067 & 0.624 & --     \\
SCAFFOLD              & \checkmark & 0.107 & 0.065 & 0.672 & \cellcolor{green!25}0.003* \\
SCAFFOLD              & --         & 0.110 & 0.066 & 0.628 & --     \\
Local                 & --         & 0.097 & 0.059 & 0.755 & --     \\
\bottomrule
\end{tabular}}
\end{table}

\subsection{Localizing the Failure}
\label{subsec:res_localize}

Both diagnostics of Section~\ref{subsec:localize} are measured on building energy.
Over the four seeds common to all sweep runs, the RMSE rises with $E$ for both schemes, from \sweepAvgQuarter{} to \sweepAvgTwo{} for FedAvg and from \sweepProxQuarter{} to \sweepProxTwo{} for FedAdam+Prox between $E = 0.25$ and $E = 2$. That is the ordinary cost of local steps. The static-input sensitivity separates the two schemes. FedAdam+Prox retains \sensBdgProxEzeroone{} at $E = 0.1$ and \sensProxEquarter{} at $E = 0.25$, and falls below $0.01$ from $E = 1$ on. FedAvg is already at $0.001$ at $E = 0.25$. Under FedAdam+Prox the sensitivity thus falls steadily with $E$, without a threshold. FedAvg is silent after a quarter of an epoch, so any threshold there lies below it.

Static-path coupling does not restore the use of the static input. With the temporal pathway running full local epochs under FedAdam+Prox, the model reaches RMSE \fsepochRmse{} with the static input against \fsepochRmseNostatic{} without, a significant contrast ($p = \fsepochP{}$), yet its sensitivity is \fsepochSens{} of the central value, as silent as under FedAvg. As for SCAFFOLD in Section~\ref{subsec:res_bdg2}, the contrast comes from the additional parameters of the static encoder and not from the entity information. Synchronizing the encoder keeps its weights global, yet the static pathway stays silent, so the loss does not sit in the encoder alone. A variant that also aggregates the temporal pathway every 50 batches reaches \fedstaticRmse{} ($p = \fedstaticP{}$), yet FedAdam+Prox at $E = 0.1$ reaches \rmseBdgProxEzeroone{} without any coupling, so that gain comes from the more frequent temporal aggregation rather than from the encoder. Coupling the encoder alone is not cheap either, at \fsepochCommShare\,\% of the per-batch volume, because it synchronizes the encoder about 13\,800 times over \fsepochRounds{} epochs, whereas per-batch FedAdam stops after about 1\,700 steps. We report static-path coupling as a diagnostic and not as a scheme to adopt.

\subsection{Communication Cost}
\label{subsec:res_comm}

The federated scheme that uses the static input fully is also the most expensive to communicate.
We quantify this cost as the volume of data transmitted per client, since latency and bandwidth are deployment-specific.
Table~\ref{tab:comm} reports the volume sent and received per client and in total for the static-input configuration, with the local training-set size as reference.
Per-batch FedAdam synchronizes after every batch rather than once per round, so it transmits 31 times as much per client as FedAvg on building energy, 35 times as much on wind power, and 3.5 times as much on the small synthetic dataset. Wind power costs more than building energy because per-batch FedAdam trains longer there, about 6\,500 against 1\,700 steps.

\begin{table}[t]
\centering
\caption{Communication cost per scheme and dataset: data sent and received per client (GB/cl.) and total (GB) over $K$ clients ($K{=}12$ synthetic, $K{=}32$ applications)}
\label{tab:comm}
\resizebox{\columnwidth}{!}{%
\begin{tabular}{lcccccc}
\toprule
 & \multicolumn{2}{c}{Synthetic} & \multicolumn{2}{c}{Building energy} & \multicolumn{2}{c}{Wind power} \\
\cmidrule(lr){2-3}\cmidrule(lr){4-5}\cmidrule(lr){6-7}
 & GB/cl. & GB & GB/cl. & GB & GB/cl. & GB \\
\midrule
Train data        & 0.0002 & 0.002 & 0.037 & 1.18 & 0.0028 & 0.090 \\
\midrule
Per-batch FedAdam & 0.109 & 1.31 & 9.46 & 302.6 & 29.39 & 940.6 \\
FedAvg            & 0.031 & 0.38 & 0.301 & 9.6 & 0.842 & 27.0 \\
FedAdam+Prox      & 0.026 & 0.32 & 0.295 & 9.4 & 0.693 & 22.2 \\
SCAFFOLD          & 0.038 & 0.45 & 0.360 & 11.5 & 1.375 & 44.0 \\
\bottomrule
\end{tabular}}
\end{table}

\subsection{Discussion}
\label{subsec:discussion}

Two observations cut across the three datasets.
First, the heterogeneity across clients is severe: no scheme with local epochs clearly beats the per-client local baseline. The largest margin is SCAFFOLD on building energy, $R^2$ $0.810$ against $0.802$ and RMSE $0.502$ against $0.513$ (Tables~\ref{tab:synthetic}, \ref{tab:bdg2} and \ref{tab:wind}). For these schemes, federating does not improve on local training.

Second, on the synthetic and the wind data, pooling more data raises accuracy only when the static input travels with it.
This is clearest on the synthetic data, where the windows are mean-centered.
Without the static input the pooled model collapses to $R^2\,0.22$, far below the local baseline at 0.95, and the static input restores it.
Wind power shows the same pattern, where the pooled static-free model fails to beat local training and only the static input pushes it past.
Building energy is the exception, because the recent load history already identifies the entity and pooling helps even without the static input.
Where the static input is the entity cue, the gains from more data are realized through the static pathway, which is exactly the pathway that FedAvg and FedAdam+Prox leave silent.

\section{Conclusion}
\label{sec:conclusion}

We asked whether a federated Temporal Fusion Transformer can learn from static, client-constant features.
The answer depends on the aggregation scheme.
Per-batch FedAdam uses the static input almost as well as central training.
Communication-efficient schemes with local epochs, FedAvg and FedAdam+Prox, extract far less from it and gain nothing significant on either application dataset.
Whether client drift or communication frequency causes the loss cannot be settled here: at the same local computation a drift-corrected SCAFFOLD keeps the static pathway responsive on wind power and gains significantly there, while on building energy it recovers nothing. Full recovery still needs per-batch FedAdam.

The effect holds across a synthetic setup and two domains, building energy and wind power, and it is not tied to the context vectors of the TFT: on the synthetic and the wind data two further architectures with different conditioning mechanisms lose most or all of their static-input sensitivity as well, and on building energy, where the static input improves the error only slightly, the GRU keeps 0.37 of 1.21.
Two diagnostics, both measured on building energy, narrow it down further. The loss is graded in the number of local steps under a proximal term and complete under FedAvg, and coupling only the static encoder per batch leaves the static pathway silent.
Per-batch FedAdam is therefore preferable whenever the static input carries entity information, at the cost of one synchronization per gradient step (Table~\ref{tab:comm}).

Two directions follow.
First, on wind power drift correction was the only lever besides communication frequency that kept the static pathway responsive, and carrying server statistics into every local step, as Mime does~\cite{karimireddy2021breaking}, is the next test.
Second, where the distribution of the static features is known, its standardization could be fixed from domain knowledge instead of aggregating client statistics. We leave this to future work, along with an evaluation under realistic latency and privacy constraints.
More broadly, client-constant inputs deserve dedicated treatment in federated training.

\bibliographystyle{IEEEtranN}
\bibliography{bibliography}

@inproceedings{walterFederatedLearningBasedForecasting2024,
  title = {Towards {{Federated Learning-Based Forecasting}} of {{Renewable Energy Production}}},
  booktitle = {2024 2nd {{International Conference}} on {{Federated Learning Technologies}} and {{Applications}} ({{FLTA}})},
  author = {Walter, Viktor and Stricker, Fabian and Wagner, Andreas and Zirpins, Christian},
  date = {2024-09-17},
  pages = {128--135},
  publisher = {IEEE},
  location = {Valencia, Spain},
  doi = {10.1109/FLTA63145.2024.10840156},
  year = {2024},
  eventtitle = {2024 2nd {{International Conference}} on {{Federated Learning Technologies}} and {{Applications}} ({{FLTA}})},
  isbn = {979-8-3503-5481-2},
  langid = {english},
}

@article{walter6346310measurement,
  title={A measurement- and {ERA5}-driven simulation framework for synthetic wind power time series with turbine aging},
  author={Walter, Viktor and Wagner, Andreas},
  journal={SSRN 6346310},
  year={2026}
}

@inproceedings{walter2025strategies,
  title={Strategies to Improve Federated Forecasts of Day-Ahead PV Power Production},
  author={Walter, Viktor and Wagner, Andreas},
  booktitle={2025 3rd International Conference on Federated Learning Technologies and Applications (FLTA)},
  pages={48--55},
  year={2025},
  organization={IEEE},
  location = {Dubrovnik, Croatia},
}

@inproceedings{mcmahan2017communication,
  title={Communication-efficient learning of deep networks from decentralized data},
  author={McMahan, Brendan and others},
  booktitle={Artificial intelligence and statistics},
  pages={1273--1282},
  year={2017},
  organization={PMLR}
}

@article{woodworth2020minibatch,
  title={Minibatch vs local sgd for heterogeneous distributed learning},
  author={Woodworth, Blake E and Patel, Kumar Kshitij and Srebro, Nati},
  journal={Advances in Neural Information Processing Systems},
  volume={33},
  pages={6281--6292},
  year={2020}
}

@article{reddi2020adaptive,
  title={Adaptive federated optimization},
  author={Reddi, Sashank and others},
  journal={arXiv preprint arXiv:2003.00295},
  year={2020}
}

@article{zhang2020adaptive,
  title={Why are Adaptive Methods Good for Attention Models?},
  author={Zhang, Jingzhao and others},
  journal={Advances in Neural Information Processing Systems},
  volume={33},
  pages={15383--15393},
  year={2020}
}

@article{lim2021temporal,
  title={Temporal fusion transformers for interpretable multi-horizon time series forecasting},
  author={Lim, Bryan and Ar{\i}k, Sercan {\"O}. and Loeff, Nicolas and Pfister, Tomas},
  journal={International Journal of Forecasting},
  volume={37},
  number={4},
  pages={1748--1764},
  year={2021},
  publisher={Elsevier}
}

@inproceedings{karimireddy2020scaffold,
  title={{SCAFFOLD}: Stochastic controlled averaging for federated learning},
  author={Karimireddy, Sai Praneeth and others},
  booktitle={International Conference on Machine Learning (ICML)},
  pages={5132--5143},
  year={2020},
  organization={PMLR}
}

@inproceedings{li2020federated,
  title={Federated optimization in heterogeneous networks},
  author={Li, Tian and others},
  booktitle={Proceedings of Machine Learning and Systems (MLSys)},
  volume={2},
  pages={429--450},
  year={2020}
}

@inproceedings{lella2025federated,
  title={Federated Learning with Temporal Fusion Transformer for Interpretable SOH Prediction in EV Batteries},
  author={Lella, Veerakotlu and others},
  booktitle={2025 IEEE 2nd International Conference on Green Industrial Electronics and Sustainable Technologies (GIEST)},
  pages={1--7},
  year={2025},
  organization={IEEE}
}

@article{ahamed2026fedtft,
  title={FedTFT: Federated Temporal Fusion Transformer for Interpretable Multi-Horizon Psychiatric Risk Prediction Across Cross-Silo Hospitals},
  author={Ahamed, Akeel and Kang, Ri-Ra and Lee, KangYoon},
  journal={IEEE Journal of Biomedical and Health Informatics},
  year={2026},
  publisher={IEEE}
}

@inproceedings{nguyen2022novel,
  title={A novel approach for federated learning with non-IID data},
  author={Nguyen, Hiep and Warrier, Harikrishna and Gupta, Yogesh},
  booktitle={2022 9th International Conference on Soft Computing \& Machine Intelligence (ISCMI)},
  pages={62--67},
  year={2022},
  organization={IEEE}
}

@article{bose2023federated,
  title={Federated Short-Term Load Forecasting with Personalization Layers for Heterogeneous Clients},
  author={Bose, Shourya and Kim, Kibaek},
  journal={arXiv preprint arXiv:2309.13194},
  year={2023}
}

@article{wang2025qkcv,
  title={{QKCV} Attention: Enhancing Time Series Forecasting with Static Categorical Embeddings for Both Lightweight and Pre-trained Foundation Models},
  author={Wang, Hao and Ma, Baojun},
  journal={arXiv preprint arXiv:2510.20222},
  year={2025}
}

@article{arivazhagan2019federated,
  title={Federated Learning with Personalization Layers},
  author={Arivazhagan, Manoj Ghuhan and Aggarwal, Vinay and Singh, Aaditya Kumar and Choudhary, Sunav},
  journal={arXiv preprint arXiv:1912.00818},
  year={2019}
}

@article{breiman2001random,
  title={Random Forests},
  author={Breiman, Leo},
  journal={Machine Learning},
  volume={45},
  number={1},
  pages={5--32},
  year={2001},
  publisher={Springer}
}

@article{miller2020building,
  title={The Building Data Genome Project 2, energy meter data from the {ASHRAE} Great Energy Predictor {III} competition},
  author={Miller, Clayton and others},
  journal={Scientific Data},
  volume={7},
  number={1},
  pages={368},
  year={2020},
  publisher={Nature Publishing Group}
}

@misc{walter2026wind,
  author    = {Walter, Viktor},
  title     = {Synthetic Simulation Data of Wind Power Time Series in {Germany}},
  year      = {2026},
  publisher = {Mendeley Data},
  version    = {V3},
  doi       = {10.17632/xv3s82rvd2.3},
  howpublished = {\url{https://data.mendeley.com/datasets/xv3s82rvd2/3}}
}

@article{montero2021principles,
  title={Principles and algorithms for forecasting groups of time series: Locality and globality},
  author={Montero-Manso, Pablo and Hyndman, Rob J},
  journal={International Journal of Forecasting},
  volume={37},
  number={4},
  pages={1632--1653},
  year={2021},
  publisher={Elsevier}
}

@article{xu2026fedps,
  title={{FedPS}: Federated Data Preprocessing via Aggregated Statistics},
  author={Xu, Xuefeng and Cormode, Graham},
  journal={arXiv preprint arXiv:2602.10870},
  year={2026}
}

@article{chen2025fadamgc,
  title={Gradient Correction in Federated Learning with Adaptive Optimization},
  author={Chen, Evan and others},
  journal={arXiv preprint arXiv:2502.02727},
  year={2025}
}

@article{hart1992nonintrusive,
  title={Nonintrusive appliance load monitoring},
  author={Hart, George W.},
  journal={Proceedings of the IEEE},
  volume={80},
  number={12},
  pages={1870--1891},
  year={1992}
}

@inproceedings{bao2024provable,
  title={Provable Benefits of Local Steps in Heterogeneous Federated Learning for Neural Networks: A Feature Learning Perspective},
  author={Bao, Yajie and Crawshaw, Michael and Liu, Mingrui},
  booktitle={International Conference on Machine Learning (ICML)},
  volume={235},
  pages={2857--2902},
  year={2024},
  organization={PMLR}
}

@article{karimireddy2021breaking,
  title={Breaking the centralized barrier for cross-device federated learning},
  author={Karimireddy, Sai Praneeth and others},
  journal={Advances in Neural Information Processing Systems},
  volume={34},
  pages={28663--28676},
  year={2021}
}

@article{zhao2018federated,
  title={Federated learning with non-{IID} data},
  author={Zhao, Yue and others},
  journal={arXiv preprint arXiv:1806.00582},
  year={2018}
}

@inproceedings{lee2023layerwise,
  title={Layer-wise adaptive model aggregation for scalable federated learning},
  author={Lee, Sunwoo and Zhang, Tuo and Avestimehr, A. Salman},
  booktitle={Proceedings of the AAAI Conference on Artificial Intelligence},
  volume={37},
  number={7},
  pages={8491--8499},
  year={2023}
}

\end{document}